%% file: main.tex
\documentclass[10pt,twocolumn,letterpaper]{article}

\usepackage{cvpr}              %
\usepackage{bm}      %
\usepackage{multirow}
\usepackage{makecell}
\usepackage{cuted}
\usepackage[accsupp]{axessibility} %
\input{preamble}

\definecolor{cvprblue}{rgb}{0.21,0.49,0.74}
\usepackage[pagebackref,breaklinks,colorlinks,citecolor=cvprblue]{hyperref}

\def\paperID{}%
\def\confName{CVPR}
\def\confYear{2024}

\title{
GeneAvatar: Generic Expression-Aware Volumetric Head Avatar Editing \\ from a Single Image
}

\author{
Chong Bao$^{1*}$\footnotemark[4]
\quad
Yinda Zhang$^{2}$\footnotemark[1] \quad
Yuan Li$^{1}$\footnotemark[1]  \quad
Xiyu Zhang$^{1}$ \quad
Bangbang Yang$^{4}$ \quad \\
Hujun Bao$^{1}$ \quad
Marc Pollefeys$^{3}$ \quad
Guofeng Zhang$^{1}$ \quad
Zhaopeng Cui$^{1}$\footnotemark[2] \\
$^{1}$State Key Lab of CAD\&CG, Zhejiang University \quad
$^{2}$Google \quad $^{3}$ETH Zürich \quad $^{4}$ByteDance\\
}

\makeatletter
\DeclareRobustCommand\onedot{\futurelet\@let@token\@onedot}
\def\@onedot{\ifx\@let@token.\else.\null\fi\xspace}

\def\eg{e.g\onedot} 

\def\ie{i.e\onedot}

\makeatother

\newcommand{\ssecspace}{\vspace{-0.5em}}

\newcommand{\secspace}{\vspace{-0.5em}}

\begin{document}
\twocolumn[{%
\renewcommand\twocolumn[1][]{#1}%
\vspace{-3.6em}
\maketitle
\vspace{-3.3em}
\begin{center}
    \centering
    \includegraphics[width=1.0\linewidth, trim={0 0 0 0}, clip]{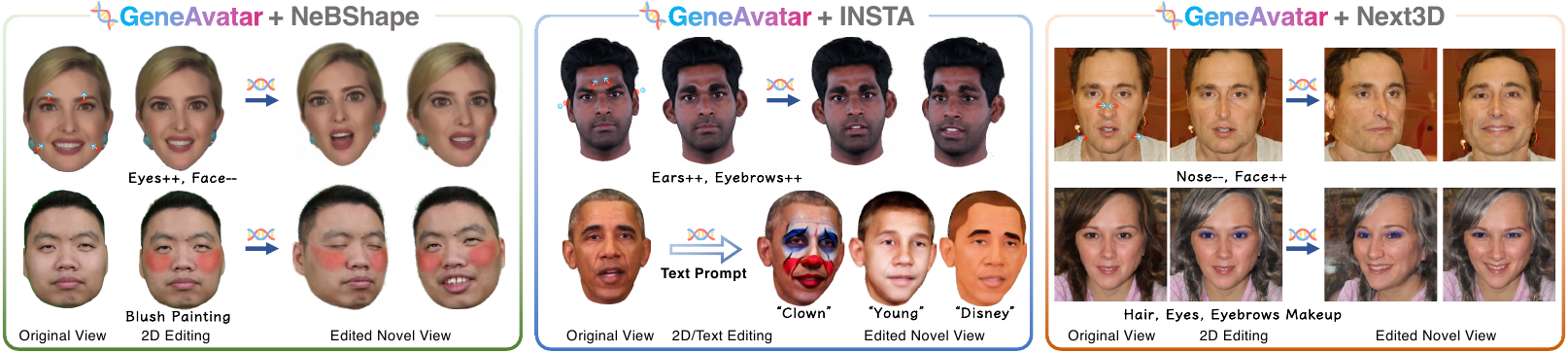}
        \vspace{-2.0em}
    \captionof{figure}{We propose a generic approach to edit 3D avatars in various volumetric representations (NeRFBlendShape~\cite{nerfblendshape}, INSTA~\cite{insta}, Next3D~\cite{sun2023next3d}) from a single perspective using 2D editing methods with drag-style, text-prompt and pattern painting. Our editing results are consistent across multiple facial expression and camera viewpoints.
    }
    \vspace{-0.5em}
    \label{fig:teaser}
\end{center}%
}]

\renewcommand{\thefootnote}{\fnsymbol{footnote}}
\footnotetext[1]{Authors contributed equally.}
\footnotetext[2]{Corresponding authors.}
\footnotetext[4]{The work was partially done when visiting ETHZ.}

\maketitle

\input{sec/0_abstract}    
\secspace
\vspace{-1.0em}
\input{sec/1_intro}

\secspace
\input{sec/2_formatting}
\input{sec/3_finalcopy}

\secspace
\input{sec/4_experiments}

\secspace
\section{Conclusion}
\secspace

We have proposed a novel generic editing approach that allows users to edit various volumetric head avatar representations from a single image, where an expression-aware modification generator lifts the editing to the 3D avatar while maintaining consistency across multiple expression and viewpoints.
As a limitation, we cannot add additional objects (\eg, hat) or modify the hairstyle
as shown in our supplementary Sec.~\textcolor{red}{C.5}, which may be improved by learning extra specialized geometry addition and hair modification generators.

\noindent\textbf{Acknowledgment:} This work was partially supported by the NSFC (No.~62102356) and Ant Group.

{
    \small
    \bibliographystyle{ieeenat_fullname}
    \bibliography{main}
}

\clearpage

\input{supplementary}

\end{document}

%% file: preamble.tex
\usepackage[dvipsnames]{xcolor}

%% file: sec/0_abstract.tex
\begin{abstract}
\vspace{-1.0em}

Recently, we have witnessed the explosive growth of various volumetric representations in modeling animatable head avatars.
However, due to the diversity of frameworks, there is no practical method to support high-level applications like 3D head avatar editing across different representations.
In this paper, we propose a generic avatar editing approach that can be universally applied to various 3DMM-driving volumetric head avatars.
To achieve this goal, we design a novel expression-aware modification generative model, which enables
lift 2D editing from a single image to a consistent 3D modification field.
To ensure the effectiveness of the generative modification process,
we develop several techniques, including an expression-dependent modification distillation scheme to draw knowledge from the large-scale head avatar 
model and 2D %
facial texture editing tools, implicit latent space guidance to enhance model convergence, and a segmentation-based loss reweight strategy for fine-grained texture inversion.
Extensive experiments demonstrate that our method delivers high-quality and consistent results across multiple expression and viewpoints.
Project page: \href{https://zju3dv.github.io/geneavatar/}{https://zju3dv.github.io/geneavatar/}.

\end{abstract}

%% file: sec/1_intro.tex
\section{Introduction}
\label{sec:intro}

Recently various volumetric representations~\cite{gafni2021dynamic, zheng2022avatar, nerfblendshape, insta, bai2023learning, athar2022rignerf, zheng2023pointavatar, xu2023avatarmav} %
have achieved remarkable success in reconstructing personalized, animatable, and photorealistic head avatars using implicit~\cite{gafni2021dynamic, zheng2022avatar, nerfblendshape, zheng2023pointavatar, xu2023avatarmav} or explicit~\cite{insta, bai2023learning, athar2022rignerf} conditioning of 3D Morphable Models (3DMM) \cite{blanz1999morphable}. 
A popular demand, once with a created avatar model, is to edit the avatar, \eg, for face shape, facial makeup, or apply artistic effects, for the downstream applications, \eg, in virtual/augmented reality. 

Ideally, the desired editing functionality on the animatable avatar should have the following properties.
(1) \textbf{Adaptable}: 
The editing method should be applicable across various volumetric avatar representations. 
This is particularly valuable in light of the growing diversity of avatar frameworks \cite{insta,nerfblendshape,sun2023next3d}.
(2) \textbf{User-friendly}: 
The editing should be user-friendly and intuitive. Preferably, 
the editing of geometry and texture of the 3D avatar could be accomplished on a single-perspective rendered image.
(3) \textbf{Faithful}: The editing results should be consistent across various facial expression and camera viewpoints.
(4) \textbf{Flexible}: 
Both intensive editing (\eg, global appearance transfer following style prompts) and delicate local editing (\eg, dragging to enlarge eyes or ears) should be supported as illustrated in Fig.~\ref{fig:teaser}.

However, 3D-aware avatar editing is still underexplored in both geometry and texture.
One plausible way is to perform 3D editing via animatable 3D GAN~\cite{sun2023next3d, wu2022anifacegan,tang20233dfaceshop},
but the editing results may not be consistently reflected when expression and camera viewpoint change.
Alternatively, the editing can be done on the generated 2D video using 2D personalized StyleGAN~\cite{lin2023pvp}; however, the identity shift is often observed.
Some face-swapping methods \cite{faceswap1, faceswap2, roop} are capable of substituting the face in a video with another face derived from a reference image or video; however, they do not support texture editing and local geometry editing.

To this end, we propose GeneAvatar -- a generic approach to 
support fine-grained 3D editing
in various volumetric avatar representations from a single perspective by leveraging 2D editing methods, such as drag-based methods~\cite{pan2023drag, ling2023freedrag, mou2023dragondiffusion, shi2023dragdiffusion}, text-driven methods~\cite{brooks2023instructpix2pix,hertz2023delta,parmar2023zero,ge2023expressive,patashnik2021styleclip}, or image editing tools like Photoshop (see Fig.~\ref{fig:teaser}).
We adopt a novel editing framework that formulates the editing as predicting expression-aware 3D modification fields applied in the geometry and texture space of the volumetric avatars, which makes editing independent with the original representation as long as they are in parametric-driven radiance field, \eg, 3DMM-based neural avatar. 
Second, to ensure that the 2D image editing can be faithfully transferred into the 3D space, we propose to learn a generative model for modification fields, which produces 3DMM conditional modification fields from a compact latent space.
Given the rendered avatar image and its edited counterpart, we conduct auto-decoding optimization on this generative model to search for the latent code that best explains the editing, obtaining consistent 3DMM conditional modification fields across various viewpoints and expression.
Third, inspired by the spirit of learning from the pre-trained large-scale generative model~\cite{haque2023instruct,mikaeili2023sked,zhuang2023dreameditor,brooks2023instructpix2pix},
we design a novel distillation scheme to learn the expression-dependent modification from a 3DMM-based GAN~\cite{sun2023next3d} and 2D face editing tools~\cite{li2018beautygan, jiang2020psgan, nguyen2021lipstick}.
The scheme addresses the issue of insufficient real training data (\ie, avatars with a wide range of geometry and texture changes).
Besides, we develop several techniques to enhance the editing effects, including the implicit latent space guidance to stabilize the initialization and convergence of learning, and a segmentation-based loss reweight strategy for fine-grained texture inversion.

The contributions of our paper are summarized as follows.
\textbf{1)} We propose a generic avatar editing approach that can be applied to various 3DMM driving head avatars in the neural radiance field.
To achieve this, we design a novel expression-aware modification generative model, which lifts the geometry and texture editing from a single image to a consistent 3D modification field.
\textbf{2)} 
To bootstrap the training of the modification generator with limited %
real paired training data,
we design a distillation scheme to learn the expression-dependent geometry and texture modification from the large-scale head avatar generative model~\cite{sun2023next3d} and 2D face texture editing tools~\cite{nguyen2021lipstick,jiang2020psgan,li2018beautygan}, and develop several techniques, including implicit guidance in latent space to improve training convergence, and a loss reweight strategy based on segmentation for fine-grained texture inversion.
\textbf{3)} Extensive experiments on various head avatar representations demonstrate that our method delivers high-quality editing results and the editing effects are consistent under different viewpoints and expression.

%% file: sec/2_formatting.tex
\section{Related Work}
\label{sec:relatedwork}
\secspace

\noindent\textbf{2D Head Avatar Editing.}
The manipulation of 2D head avatars has made significant strides in recent years.
Various GAN-based methods~\cite{stylegan, stylegan2, karras2021alias} can result in precise and high-resolution human face editing by leveraging image space semantic information~\cite{zhu2020sean, editgan} or controlling latent space explorations~\cite{zhu2020domain, harkonen2020ganspace, cherepkov2021navigating, shen2020interpreting}.
Some approaches~\cite{li2018beautygan, yang2022elegant,jiang2020psgan, nguyen2021lipstick} focus on the task of makeup transferring by exploiting the GAN to learn the transferring ability from a large unaligned makeup and non-makeup face datasets.
The drag-based GAN editing approach~\cite{zhu2023linkgan,pan2023drag} gained vast popularity due to providing a user-friendly editing way.
PVP~\cite{lin2023pvp} uses a monocular video to fine-tune StyleGAN~\cite{stylegan, stylegan2} to obtain the personalized image generator and provide various editing functions.
The diffusion models~\cite{brooks2023instructpix2pix, huang2023composer} also show the capability of achieving fine-grained face editing with text prompt and other conditional input.
For editing an avatar in a video, lots of face-swapping methods~\cite{faceswap1, faceswap2, roop} have emerged to provide high-quality and properly aligned face-swapping results.
However, these approaches typically suffer from multi-view consistency and identity preservation.

\noindent\textbf{3D Head Avatar Editing.}
Neural Radiance Field~\cite{nerf} has exhibited great reconstruction and rendering qualities in SLAM~\cite{zhu2022nice, yang2022vox}, scene editing~\cite{neumesh, bao2023sine, object_nerf, yang2022neural, zhang2024coarf} and relighting~\cite{zhang2021nerfactor, ye2023intrinsicnerf, zhao2022factorized}, especially promoting the emergence of many 3D avatar reconstruction~\cite{insta,nerfblendshape,zheng2022avatar,zheng2023pointavatar,bai2023learning,wang2021learning} and generation~\cite{sun2023next3d,wu2022anifacegan,tang20233dfaceshop}.
Some methods~\cite{zhang2023fdnerf, sun2022ide, abdal20233davatargan, xu20223d} exploit the powerful editing ability of GAN to edit a 3D static head portrait. However, they cannot be trivially extended to the dynamic avatars.
The methods~\cite{pan2023avatarstudio, shao2023control4d, nguyen2023alteredavatar} focus on style transfer of the avatar using text prompt or style image but reach a poor identity-preserving.
We propose a novel 3D avatar editing approach with an expression-aware modification generative model, which can be applied to various 3DMM-based volumetric avatars and render consistent novel views with fine-grained editing across multiple viewpoints and expression while preserving identity of person.

%% file: sec/3_finalcopy.tex
\begin{figure*}[!t]
\centering
\vspace{-1.7em}
\includegraphics[width=0.97\linewidth, trim={0 0 0 0}, clip]{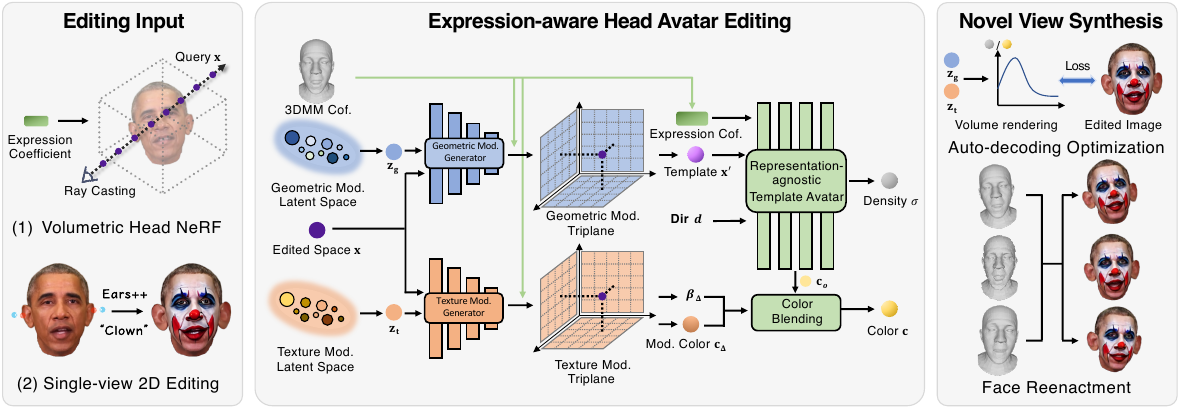}
\vspace{-1.0em}
\caption{We use an expression-aware generative model that accepts a modification latent code $\mathbf{z}_{g/t}$ and 3DMM coefficients and outputs a modification field of a tri-plane structure.
The modification field modifies the geometry and texture of the template avatar by deforming the sample points $\mathbf{x}$ and blending the color $\mathbf{c}_{o}$ with the modification color $\mathbf{c}_{\Delta}$ respectively.
We lift the 2D editing effect to 3D using an auto-decoding optimization and synthesize novel views across different expression.
\vspace{-1.5em}
    }
\label{fig:framework}
\end{figure*}

\vspace{-1.7em}
\section{Method}
As shown in Fig.~\ref{fig:framework}, given a volumetric head avatar, we edit the avatar using a single-view image and synthesize consistent novel views across multiple expression and viewpoints.
To achieve this goal, we propose a novel expression-aware modification generator to generate 3D modification fields, which can be seamlessly integrated into various representations and animated with facial expression (see Sec.~\ref{ssec:generator}).
Furthermore,
to bootstrap the training with limited pairwise data,
we propose a novel expression-aware distillation scheme to learn the expression-dependent modifications from large-scale generative models~\cite{sun2023next3d, brooks2023instructpix2pix} (see Sec.~\ref{ssec:learning}).
During the editing process,
given a single edited image of a 3D avatar, we perform an auto-decoding optimization to lift 2D editing effect to the 3D space~(see Sec.~\ref{ssec:avatar_editing}).

\subsection{Preliminaries}
\label{ssec:implicit_rep}
\ssecspace

The current implicit volumetric representations of head avatar~\cite{nerfblendshape, insta, sun2023next3d} are mostly built upon the NeRF~\cite{nerf} or its variants~\cite{nsvf, yu2021plenoxels, sun2022direct, chen2022tensorf, eg3d, muller2022instant}. In general, 
the neural architecture 
can be simplified as an implicit field $\mathbf{F}$ that takes position $\mathbf{x}$ and view direction $\mathbf{d}$ as inputs and predicts the geometry $\sigma$ and the texture $\mathbf{c}$ of the avatar, \ie, $(\sigma, \mathbf{c}) = \mathbf{F}(\mathbf{x}, \mathbf{d})$.
Then, the volume rendering technique is used to render images as follows: %
\begin{equation}
\begin{split}
        & \hat{C}(\bm{r}) = \sum_{i=1}^{N} T_i \alpha_i {\mathbf{c}}_i, \;\;\; 
        T_i = \exp{\left(-\sum_{j=1}^{i-1}{\sigma'}_j \delta_j\right)},
\label{eq:rendering}
\end{split}
\end{equation}
where $\alpha_i = 1-\exp{(-{{\sigma}'}_i \delta_i)}$, and $\delta_i$ is the distance between adjacent samples along the ray. 
In order to animate the head avatar, 3DMM \cite{blanz1999morphable} is incorporated to describe the deformation implicitly~\cite{gafni2021dynamic, zheng2022avatar, nerfblendshape, zheng2023pointavatar, xu2023avatarmav} or explicitly~\cite{insta, bai2023learning}.

\subsection{Expression-aware Modification Generator}
\label{ssec:generator}
\ssecspace

To enable the modification animated with the facial expressions,
we follow the architecture of 3DMM-based 3D GAN~\cite{sun2023next3d} to build our expression-aware modification generator. As shown in Fig.~\ref{fig:framework}, 
our generator consists of a geometry generator $\mathbf{G}_{{\Delta}g}$ and a texture generator $\mathbf{G}_{{\Delta}t}$.
$\mathbf{G}_{{\Delta}g}$ encodes the expression-dependent geometry modification by deforming the query points in the edited space to the original template space under each expression.
$\mathbf{G}_{{\Delta}t}$ encodes the expression-dependent modification color of query points under each expression:
\begin{equation}
\label{eq:network}
    \mathbf{x}' = \mathbf{G}_{{\Delta}g}(\mathbf{x}, \mathbf{z}_g, \mathbf{v}),\; (\mathbf{c}_{\Delta}, \beta_{\Delta}) = \mathbf{G}_{{\Delta}t}(\mathbf{x}, \mathbf{z}_t, \mathbf{v}),
\end{equation}
where $\mathbf{x}$ is the query points in the edited space, and $\mathbf{x}'$ is the deformed point in the space of original avatar under current expression $\mathbf{e}$. 
$\mathbf{c}_{\Delta}$ is the modification color and $\beta_{\Delta}$ determines the blending weights with the original color.
$\mathbf{z}_g, \mathbf{z}_t$ are the geometry and texture modification latent code respectively, where $\mathbf{z}_g, \mathbf{z}_t \in R^{1024}$.
They control the generation of modification feature maps in the UV space.
$\mathbf{v}$ is the 3DMM mesh vertices~\cite{flame} that condition the current expression $\mathbf{e}$.
Each mesh vertex has a neural feature that is retrieved in the modification feature map using pre-defined UV mapping.
We rasterize the vertex features to the three axis-aligned planes to generate the tri-plane feature.
The modification information of the query point $\mathbf{x}$ is first collected by bilinear interpolation on the tri-plane feature and then decoded by the neural feature decoder~\cite{eg3d}.
Since the modification is defined as decoupled fields without relying on the original field, our generated modification field can be integrated into various volumetric avatar representations and be animated following the facial expression.

\ssecspace
\subsection{Expression-dependent Modification Learning}
\label{ssec:learning}
\ssecspace
To learn the proposed expression-aware modification, we need extensive training data on avatars with a wide range of geometry and texture changes, which is hard to obtain in practice.
Following the spirit of learning high-fidelity editing ability from the large-scale generative model~\cite{haque2023instruct,mikaeili2023sked,zhuang2023dreameditor,brooks2023instructpix2pix}, we propose a novel expression-aware distillation scheme to deal with %
insufficient real training data.
We leverage the ability of 3DMM-based 3D GAN~\cite{sun2023next3d} and 2D face texture editing tools to generate facial editing data, which encompasses a wide range of geometry and texture editing across various expression and viewpoints.

\noindent\textbf{Geometry Distillation.}
We use the teacher 3DMM-based 3D GAN~\cite{sun2023next3d} $\mathbf{G}_n$ to synthesize two volumetric avatars with different geometry (an original avatar $\mathbf{F}$ and an edited avatar $\mathbf{F}'$) by modifying the 3DMM shape parameter of the original avatar.
This provides the paired editing data for our generator to learn how to modify the geometry of the avatar while maintaining consistency across various expression and viewpoints.
Specifically, we randomly sample the latent code $\mathbf{z}_n$ in the latent space of $\mathbf{G}_n$ as well as 3DMM shape parameter $\bm{\beta}$, expression parameter $\bm{\psi}$, and pose parameter $\bm{\theta}$.
$\bm{\beta}, \bm{\psi}$ are sampled from
a normal distribution whose absolute mean and standard deviation are within $[0,1]$.
$\bm{\theta}$ are a group of rotation vectors that have random directions within a unit sphere and magnitude within $[-6, 6]$ degrees.
Then, we sample an edit vector $\bm{\beta}_{\Delta}$ from a uniform distribution $\mathcal{U}(-3, 3)$
and apply it to the original shape parameter by $\bm{\beta}' = \bm{\beta}_{\Delta} + \bm{\beta}$.
These hyperparameters w.r.t. 3DMM coefficients sampling are selected empirically to maintain the shape definition of the human head.
Please refer to our supplementary Sec.~\textcolor{red}{B.2} for more details on 3DMM sampling.
The original avatar $\mathbf{F}$ and paired edited avatar $\mathbf{F}'$ are generated by $\mathbf{F} = \mathbf{G}_n(\mathbf{z}_n, \bm{\beta}, \bm{\psi}, \bm{\theta}), \mathbf{F}' = \mathbf{G} (\mathbf{z}_n, \bm{\beta}', \bm{\psi}, \bm{\theta})$.
During training, we will apply our modification generator to modify the geometry of F such that $F$ and $F'$ render the face with the same geometry.

\noindent\textbf{Texture Distillation.}
We distill the capabilities of fine-grained texture editing from 2D face editing algorithms by generating texture-modified avatar $\mathbf{F'}$
with the teacher 3DMM GAN~\cite{sun2023next3d} $\mathbf{G}_{n}$.
Specifically, we sample an original avatar $\mathbf{F} = \mathbf{G}_n(\mathbf{z}_n, \bm{\beta}, \bm{\psi}, \bm{\theta})$ from the teacher generator and render the image of its positive face.
A segmentation-based 2D face texture editing algorithm (SBA)~\cite{face-makeup-pytorch} and two makeup transfer algorithms (MTA)~\cite{jiang2020psgan, li2018beautygan, nguyen2021lipstick} are referred to in the distillation. 
We randomly choose one of them to edit the texture of the rendered face image.
For SBA, we define several editable semantic regions of the face.
A subset of these regions is selected randomly for texture painting using hues randomly sampled from the HSV color spectrum.
For MTA, we randomly choose a makeup image as a reference from the open-sourced makeup dataset~\cite{nguyen2021lipstick,jiang2020psgan,li2018beautygan} and transfer the reference makeup to the rendered face image.
The makeup dataset~\cite{nguyen2021lipstick} contains complex makeups, such as blushes and makeup jewelry, which allow our generator to learn complicated texture editing patterns.
Then, we perform the PTI inversion~\cite{roich2022pivotal} on the texture-modified face image to lift the 2D texture editing to 3D space and obtain a texture-modified avatar $\mathbf{F'}$.

\noindent\textbf{Modification Learning.}
Following the training style of StyleGAN~\cite{stylegan2}, we sample a modification latent code $\mathbf{z}_{g/t} \in R^{1024}$ in $\mathcal{Z}$ latent space for each paired editing data.
We do not fully sample a 1024-dimensional modification code but sample a reduced code $\bar{\mathbf{z}}$ from a standard normal distribution and concatenate it with the latent code $\mathbf{z}_n$ of the original avatar $\mathbf{F}$ that is sampled from the teacher model, \ie, $\mathbf{z}_{g/t} = (\mathbf{z}_n, \bar{\mathbf{z}}), \mathbf{z}_n, \bar{\mathbf{z}} \in R^{512}$.
This design is regarded as implicit code guidance that decently integrates knowledge from the teacher model to facilitate model convergence.
$\mathbf{z}_n$ encodes the facial appearances of avatar $\mathbf{F}$, serving as a reference to the superimposition of the modification onto the avatar $\mathbf{F}$.
Note that during inference, we do not require the concatenation of latent code from the teacher model and directly optimize the full modification code from the edited image using an auto-decoding manner.
Our generator generates the modification field following Eq.~(\ref{eq:network}) where 
$\mathbf{v}$ is decoded from the 3DMM parameters of the avatar $\mathbf{F}$ using FLAME model~\cite{flame} $\mathbf{E}$,
\ie, $\mathbf{v} = \mathbf{E}(\bm{\beta}, \bm{\psi}, \bm{\theta})$.
To apply the modification field, we feed the deformed query points $\mathbf{x}'$ to the original avatar $\mathbf{F}$ to obtain the density and color, and composite the color with modification color by:
\begin{equation}
\label{eq:blending}
    \mathbf{c} = (1 - \beta_{\Delta}) * \mathbf{c}_o + \beta_{\Delta} * \mathbf{c}_{\Delta},\;\; (\sigma, \mathbf{c}_o) = \mathbf{F}(\mathbf{x}', \mathbf{d}).
\end{equation}
Then, we perform volume rendering on the density $\sigma$ and color $\mathbf{c}$ using Eq.~(\ref{eq:rendering}) to render the modified image $\hat{I}_e$ of avatar $\mathbf{F}$.
We use the photometric loss to supervise the modified image with the rendered image $\hat{I}'$ from the edited avatar $\mathbf{F}'$ under the same camera parameters.
\begin{equation}
\label{eq:blending}
    \mathcal{L} = ||\hat{I}_e - \hat{I}'||^2_2.
\end{equation}
During training, we sample multiple viewpoints and 3DMM expression parameters $\bm{\psi}$ for each editing pair $(\mathbf{F}, \mathbf{F}')$ to enhance the spatial consistency under different expressions.

\ssecspace
\subsection{Avatar Editing with Single Image}
\label{ssec:avatar_editing}
\ssecspace
In our task, users are allowed to edit a single image with various out-of-box face editing tools, such as Photoshop, drag-based editing~\cite{pan2023drag, ling2023freedrag}, text-driven editing~\cite{brooks2023instructpix2pix}.
For each editing input, we use the auto-decoding optimization on modification code to lift 2D edits into a 3D expression-aware modification field generated by our model.
This field adapts to expression and viewpoint changes and is not tied to the specific avatar representation.
The StyleGAN-based generator~\cite{stylegan2,roich2022pivotal, xia2022gan} features a latent space mapping from $\mathbf{z} \in R^{Z}$ in $\mathcal{Z}$ to $\mathbf{w} \in R^{W_n \times W_d}$ in $\mathcal{W}$, where $\mathbf{w}$ is more influential as it conditions the generator.
Therefore, we perform code inversion in $\mathcal{W}$ space by randomly sampling a modification code $\mathbf{w}_{g/t}$ during editing. This code conditions a modification field $\mathbf{G}_{{\Delta}g/t}(\mathbf{x}, \mathbf{w}_{g/t}, \mathbf{v})$ following Eq.~(\ref{eq:network}).
We apply the modification field to the original avatar using Eq. (\ref{eq:blending}). The modified image $\hat{I}_e$ is rendered following the original avatar's rendering pipeline and is encouraged to match the user-edited image $I_e$ by optimizing $\mathbf{w}_{g/t}$ with the following loss terms:
\begin{equation}
\label{eq:editing}
    \mathcal{L}_i = \lambda_1 \mathcal{L}_2(\hat{I}_e, I_e) + \lambda_2 \mathcal{L}_{\text{lpips}}(\hat{I}_e, I_e) + \lambda_3 \mathcal{L}_{\text{reg}}(\mathbf{w}, \mathbf{w}_{\text{avg}}),
\end{equation}
The L2 loss term $\mathcal{L}_2$ and LPIPS perceptual loss term $\mathcal{L}_{\text{lpips}}$ encourage the rendered face close to the appearance and structure of the edited face.
We set $\lambda_1=1000, \lambda_2=1, \lambda_3=1$.
The regularization term $\mathcal{L}_{\text{reg}}$ applied to the latent code $\mathbf{w}$ enforces alignment with the distribution of the $\mathcal{W}$ space, with $\mathbf{w}_{\text{avg}}$ representing the mean latent code computed from 1000 random samples within the $\mathcal{W}$ space.
Besides, we observe that the L2 loss on the whole face will give an underfitting result for the fine-grained makeup on the eyes, eyebrows and lips.
Therefore, we reweight the L2 loss on the facial features by face segmentation mask $M = \{N_i | i = 1,...m\}$ for texture editing: 
\begin{equation}
\label{eq:segment_loss}
\mathcal{L}_2=\sum_{i=1}^{m}  \sum_{r \in N_i}  \frac{1}{|N_i|} ||\hat{C}_e(r) - C_e(r)||_2^2,
\end{equation}
where $\hat{C}_e(r), C_e(r)$ are the rendered and target color respectively, $N_i$ are the rays within the $i$-th semantic part of the face.
Generally, we freeze the weight of the modification generator and only optimize the modification latent code $\mathbf{w}$ to reach a satisfied 3D modification result.
When an intense makeup or complicated pattern is painted onto the human face, we will continuously fine-tune the weight of our generator and freeze the latent code $\mathbf{w}$ to
achieve more accurate editing results.
To animate the edited avatar, users can input the new 3DMM expression parameter to the original avatar and our modification generator simultaneously.
In this way, the generated modification field tightly sticks to the original avatar and presents reasonable editing results under different expression and viewpoints.

%% file: sec/4_experiments.tex
\section{Experiments}
\label{sec:experiments}
In this section, we evaluate our avatar editing capability from a single perspective.
One major difference with static NeRF editing is that we focus on showing how the edits are correctly lifted to 3D avatars under various expression and camera viewpoints.

\begin{figure}[!t]
\centering
\vspace{-2.0em}
\includegraphics[width=0.97\linewidth, trim={0 0 0 0}, clip]{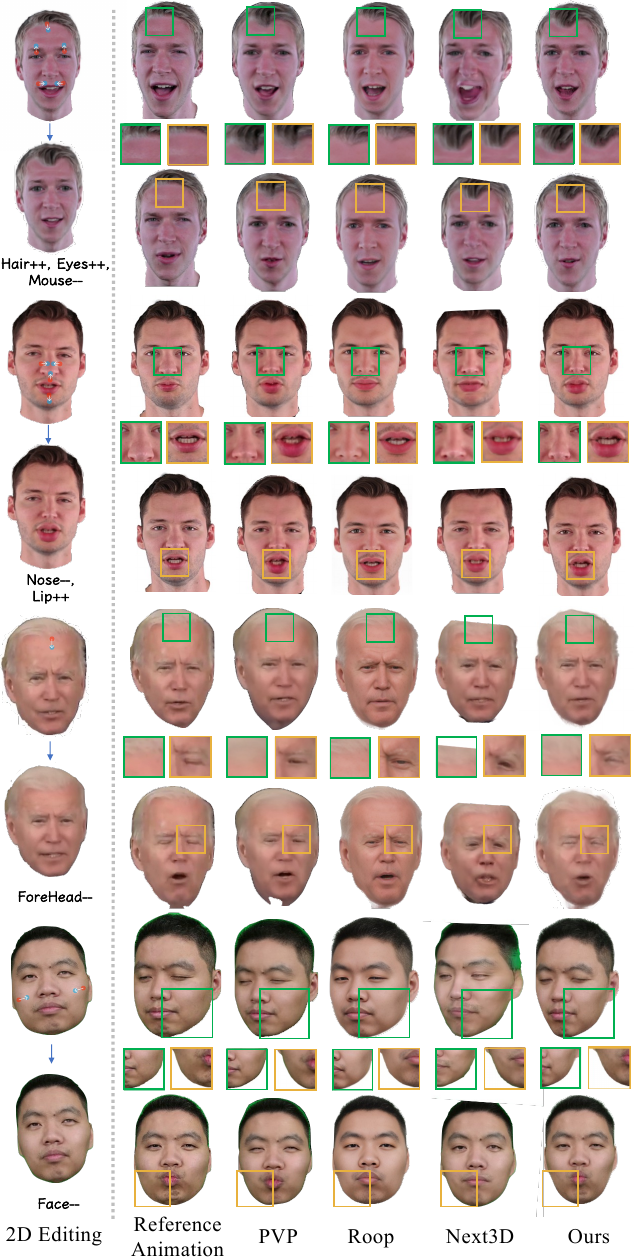}
\vspace{-1.0em}
\caption{We compare geometry editing with PVP~\cite{lin2023pvp}, Roop~\cite{roop}, Next3D~\cite{sun2023next3d} on INSTA~\cite{insta} and NeRFBlendshape~\cite{nerfblendshape} avatars. The "Reference Animation" denotes the image of the original avatar under the same expression with the rendered edited view. 
    }
\vspace{-1.5em}
\label{fig:comparisons_geo}
\end{figure}
\input{table/table_quantative_geotex}

\begin{figure*}[!t]
\centering
\vspace{-2.0em}
\includegraphics[width=0.97\linewidth, trim={0 0 0 0}, clip]{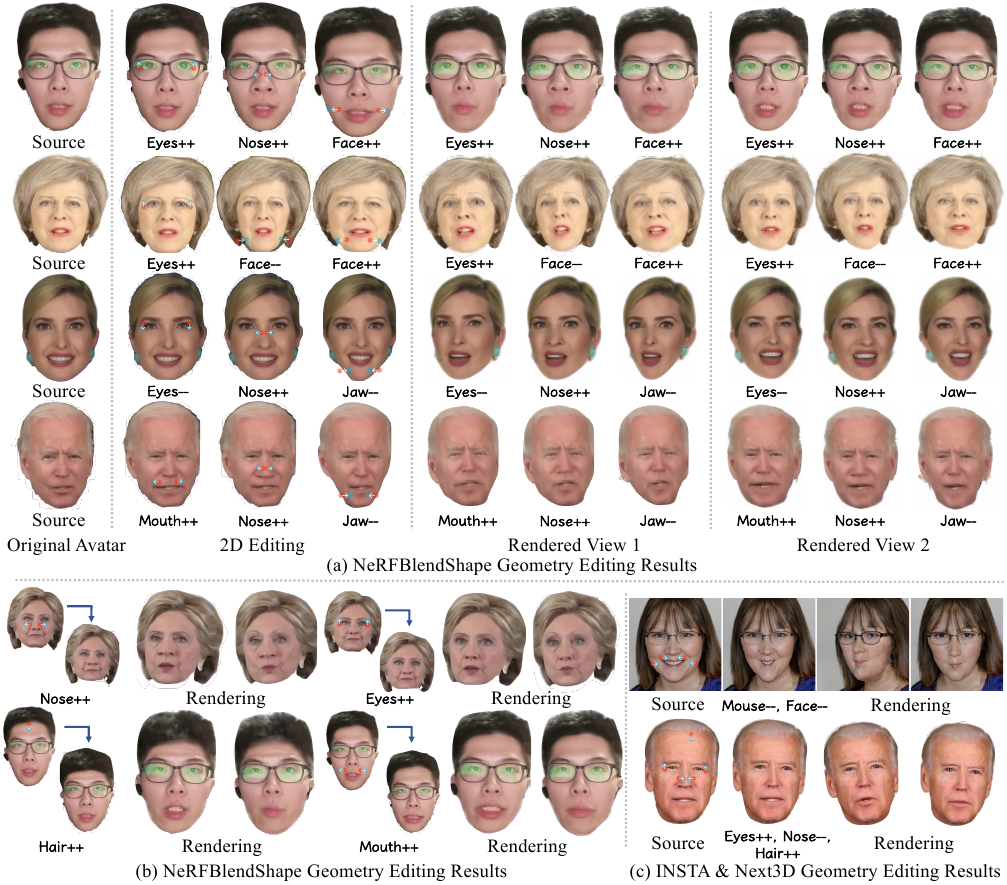}
\vspace{-1.0em}
\caption{Our geometry editing results with the drag-style 2D editing %
on INSTA~\cite{insta}, NeRFBlendshape~\cite{nerfblendshape},  and Next3D~\cite{sun2023next3d} avatars.}
\label{fig:geo_edit}
\vspace{-1.5em}
\end{figure*}

\begin{figure*}[!t]
\centering
\includegraphics[width=0.95\linewidth, trim={0 0 0 0}, clip]{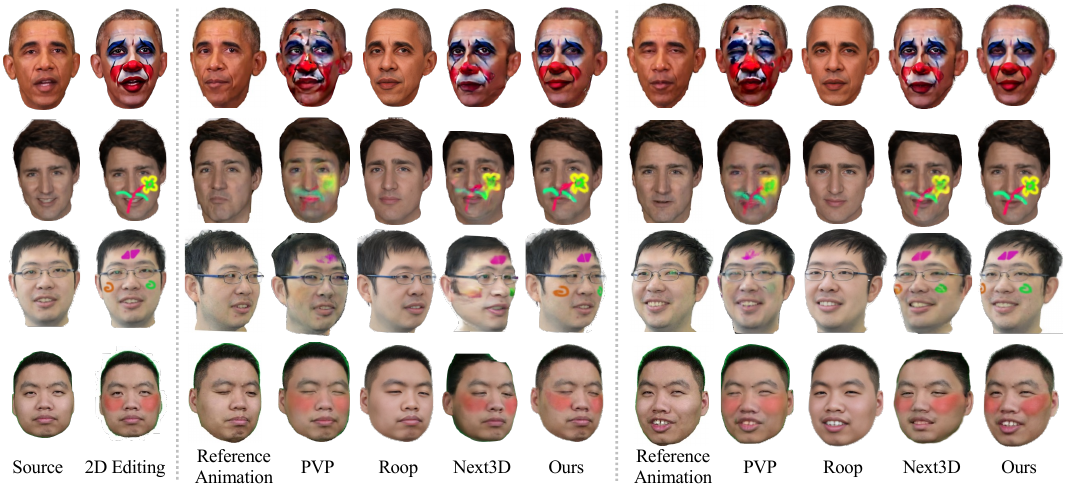}
\vspace{-1.0em}
\caption{We compare texture editing with PVP~\cite{lin2023pvp}, Roop~\cite{roop}, Next3D~\cite{sun2023next3d} on INSTA~\cite{insta} and NeRFBlendshape~\cite{nerfblendshape} avatars. The "Reference Animation" denotes the image of the original avatar under the same expression with the rendered edited view. 
    }
\label{fig:comparisons_tex}
\vspace{-1.0em}
\end{figure*}

\begin{figure}[!t]
\centering
\includegraphics[width=0.97\linewidth, trim={0 0 0 0}, clip]{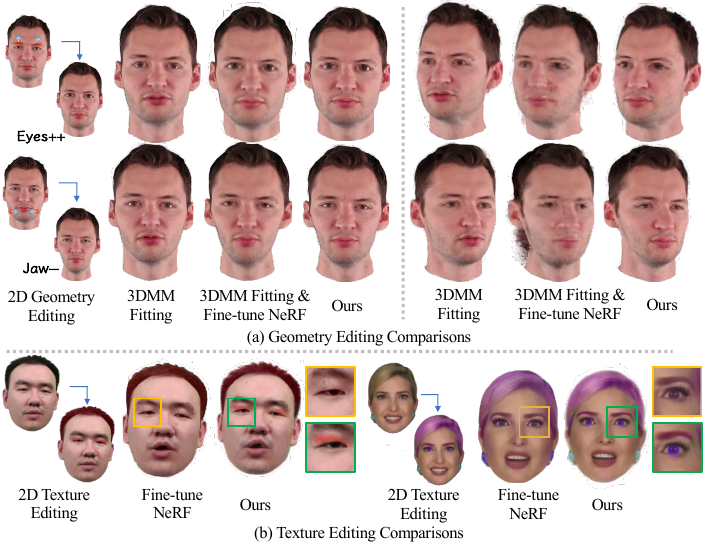}
\vspace{-1.0em}
\caption{Analysis of our effectiveness with the %
na\"ive baselines that can accomplish the single-view avatar editing.
}
\label{fig:ablation}
\vspace{-1.0em}
\end{figure}

\begin{figure*}[!t]
\centering
\includegraphics[width=0.95\linewidth, trim={0 0 0 0}, clip]{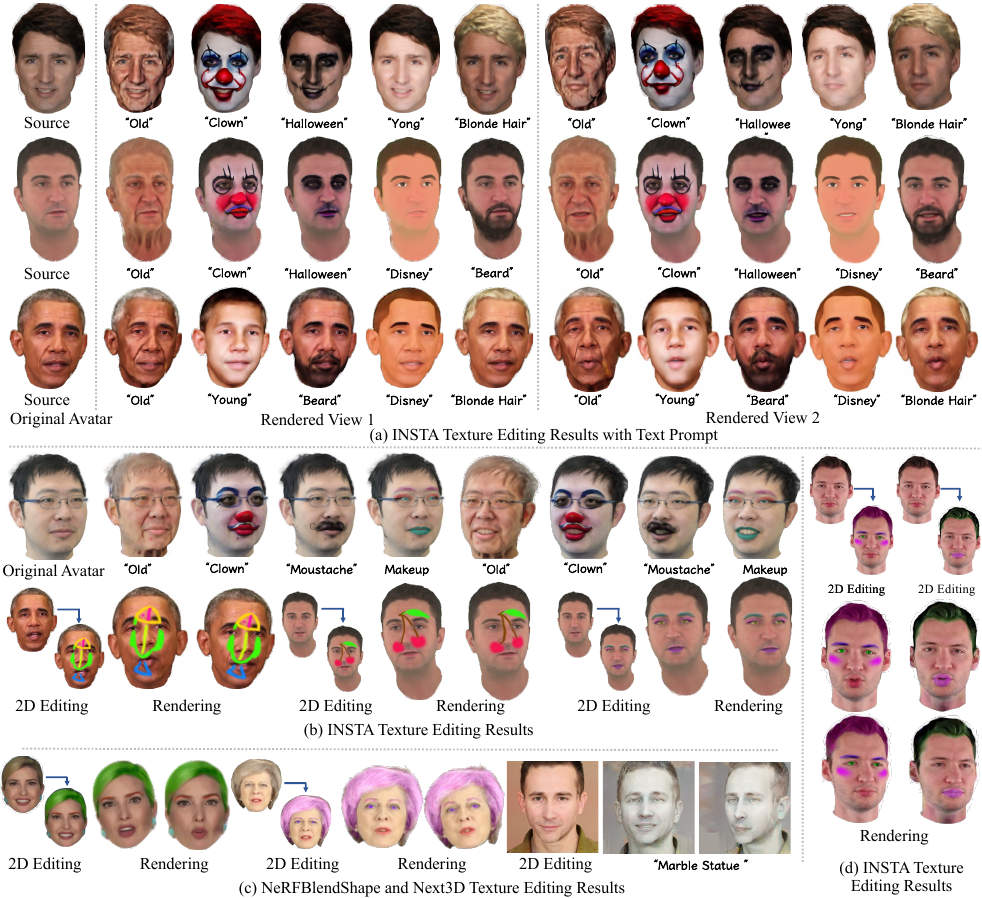}
\vspace{-0.5em}
\caption{We show our texture editing results using the 2D editing method with text-prompt, pattern painting and makeup drawing on INSTA~\cite{insta} and NeRFBlendshape~\cite{nerfblendshape}, Next3D~\cite{sun2023next3d} avatars.}
\label{fig:tex_edit}
\vspace{-1.0em}
\end{figure*}

\ssecspace
\subsection{Dataset and Baselines}
\ssecspace
\noindent\textbf{Datasets.}
We use a total of 19 neural implicit head avatars from three methods, \ie, 7 from INSTA~\cite{insta}, 8 from NeRFBlendShape~\cite{nerfblendshape}, 4 from Next3D~\cite{sun2023next3d}, and show editing results on them.
For INSTA~\cite{insta} and NeRFBlendShape~\cite{nerfblendshape}, we use the human head data (\ie, a monocular video of a head) provided by their methods to reconstruct the volumetric avatar using their respective representations.
For Next3D~\cite{sun2023next3d}, we random sample its latent space to generate volumetric avatars and perform editing on them.
The evaluation datasets exhibit a substantial variation in identities, encompassing a diverse range of races, ages, and genders.

\noindent\textbf{Baselines.} 
We pick several baseline methods that can support single-view-based avatar editing.
Roop~\cite{roop} is a face-swapping method that can swap the human face in a video from a single reference view. 
To compare with Roop, we generate videos of the original avatar rendered in driving signals and single edited frames, and perform face swap.
PVP~\cite{lin2023pvp} learns a personalized avatar image generator from a monocular video by fine-tuning the latent space of StyleGAN~\cite{stylegan2}, and performs GAN-inversion style optimization~\cite{patashnik2021styleclip} to edit the shape and appearance of the avatar.
Next3D~\cite{sun2023next3d} is a 3DMM-based 3D GAN.
To make a fair comparison on the same input data (\ie, a monocular video of avatar and an edited image), we perform GAN inversion~\cite{roich2022pivotal} with Next3D twice.
First, we fine-tune Next3D with the input video to make it learn the original geometry and texture of the avatar.
Second, we fine-tune Next3D on the edited image based on the weights of the code and generator from the first fine-tuning.

\ssecspace
\subsection{Qualitative Comparison}
\ssecspace
\noindent\textbf{Geometry Editing.} 
We first compare our method with baselines on geometry editing, \eg, changing the size of the eyes or the contour of the cheek. 
We use 2D editing tools, \eg, Photoshop and DragGAN~\cite{pan2023drag}, to modify the shape of various facial features.
Figure.~\ref{fig:comparisons_geo} shows qualitative results on avatars from INSTA~\cite{insta} (the top two) and NeRFBlendshape~\cite{nerfblendshape} (the bottom two).
Roop~\cite{roop} fails to handle the find-grained geometry change, like hairlines, and lips.
Nex3D~\cite{sun2023next3d} is able to successfully update the avatar based on the editing, however, changes the untouched part and causes an obvious identity shift.
PVP~\cite{lin2023pvp} can make edits while preserving the identity, however, the magnitude of change tends to be smaller than the given image.
In contrast, our method produces the desired editing effect from the edited image and preserves the multiview consistency and identity of the original face.

We further show more geometry editing results of our method on avatars in various representations in Fig.~\ref{fig:geo_edit}.
Our method supports a convenient way to adjust the size of diverse facial features, such as eyes, mouths, jaw, etc., by editing a single rendered image from the avatars.
The edits on 2D images are successfully lifted onto the avatar and rendered across different viewpoints and expression.
Please refer to the supplementary Sec.~\textcolor{red}{C.3} for detailed visualizations of modified head geometry.

\noindent\textbf{Texutre Editing.}
We then show our capability in texture editing.
We utilize Photoshop, an online makeup app WebBeauty ~\cite{WebAR}, and text-driven editing method Instructpix2pix~\cite{brooks2023instructpix2pix} to modify the texture on 2D renderings.
We show comparisons with PVP~\cite{lin2023pvp}, Roop~\cite{roop} and Next3D~\cite{sun2023next3d} on four distinct heads avatars (INSTA~\cite{insta} for the upper three and NeRFBlendshape~\cite{nerfblendshape} for bottom one) in Fig.~\ref{fig:comparisons_tex}. 
Roop~\cite{roop} is ineffective in transferring non-human-face-like texture, thus failing in all examples.
PVP~\cite{lin2023pvp} only transfers partial or blurry textures, and also causes shifts across the expression and head poses.
Next3D~\cite{sun2023next3d} successfully uplifts the texture editing in 2D images sharply. However, it still suffers from the identity shift issue
in the lower two heads and a blurred pattern in the upper two heads in Fig.~\ref{fig:comparisons_tex}.
In contrast, our method faithfully paints the complicated texture following the edited image and preserves the identity of the original avatar and consistency across multiple viewpoints and expressions.

We show extensive texture editing results in three avatar representations in Fig.~\ref{fig:tex_edit}.
Our method supports a wide range of texture editing, including global style transfer (``Add a clown makeup'' via text-driven editing), semantic-driven editing (changing the hair color), and free-form sketch (painting on the face).
Please refer to the supplementary Sec.~\textcolor{red}{C.1/2} for hybrid editing and face reenactment results.

\vspace{-0.5em}
\subsection{Quantitative Comparison}
\ssecspace

\noindent\textbf{User Study.} We conducted a user study to validate our method further quantitatively.
Following the evaluation protocol of Avatarstudio~\cite{pan2023avatarstudio}, users are required to watch the rendered videos of different methods side by side and answer each question by picking up one of the methods.
For each group of editing results, we will ask four questions the same as Avatarstudio~\cite{pan2023avatarstudio} on editing preservation, identity preservation, temporal consistency and overall performance in Tab.~\ref{tab:use_study}.
We collected statistics from 30 participants in 12 groups of edit results. Results are reported in Tab.~\ref{tab:use_study}. 
Our method exhibits the best editing ability while keeping the best consistency across different facial expressions for geometry and texture editing. 
Moreover, our method performs the best in keeping non-edited parts untouched and making the human heads still recognizable after being edited \eg, identity preservation metric in Tab.~\ref{tab:use_study}.
Please refer to the supplementary Sec.~\textcolor{red}{B.4} for more details.

\noindent\textbf{Image Identity Similarity Evaluation.}
Following the VoLux-GAN~\cite{tan2022volux}, we further evaluate the cross-view identity consistency in our edit results.
We take 7 groups of geometry editing results and 7 groups of texture editing results, and render the 350 images of edited avatars with different viewpoints and expressions.
We calculate the cosine similarities between each rendered image and the single-view 2D edited image and average the similarities on all rendered images as metrics.
As reported in Tab.~\ref{tab:use_study}, our method outperforms all baselines in recovering desired editing effect and retaining identity consistency.

\ssecspace
\subsection{Native Editing Capability of Avatar}
\label{sec:ablation_study}
\ssecspace
In this section, we analyze the editing capability from the original avatar model and show the necessity of our design.

\noindent \textbf{Comparison to 3DMM-based Geometry Editing.}
One intuitive way to support geometry editing is updating the underlying 3DMM geometry~\cite{feng2021learning}. Here, we investigate this method and verify its capability.
Specifically, we run state-of-the-art single image-based 3DMM reconstruction method~\cite{zielonka2022towards}, and update the 3DMM shape parameter of the avatar model with the estimated one.
Note that such an approach is only available for those avatars that use 3DMM explicitly, and we take INSTA \cite{insta} as an example and show the results in Fig.~\ref{fig:ablation} (a).
The 3DMM fitting tends to fail when the edited face is out-of-distribution, so the magnitude of editing may not be correct (\eg the face shape).
Changing the 3DMM parameter could also result in blurry rendering from the pre-trained avatar model, which cannot be trivially fixed even after fine-tuning the rendering decoder.

\vspace{-0.2em}
\noindent \textbf{Comparison to Fine-tuning for Texture Editing.}
Texture editing could be done by fine-tuning the avatar with the one-shot edited image.
We test this method on avatars from NeRFBlendShape~\cite{nerfblendshape} and show the results in Fig.~\ref{fig:ablation} (b).
While large structured changes, \eg, hair color, can be edited, detailed editing is largely ignored.
Please refer to the supplementary Sec.~\textcolor{red}{C.4} for more ablation studies.

%% file: table/table_quantative_geotex.tex
\begin{table}[tb]
\centering
\vspace{-1.5em}
\resizebox{1.0\linewidth}{!}{
\tabcolsep 3pt
\begin{tabular}{lrrrrrrrr}
\toprule
\multicolumn{1}{c}{\multirow{2}{*}{Methods}} & \multicolumn{4}{c}{Geometry} & \multicolumn{4}{c}{Texture} \\ \cmidrule(lr){2-5} \cmidrule(lr){6-9}  
\multicolumn{1}{c}{} & \multicolumn{1}{l}{Roop} & \multicolumn{1}{l}{PVP} & \multicolumn{1}{l}{Next3D} & \multicolumn{1}{l}{Ours} &  \multicolumn{1}{l}{Roop} & \multicolumn{1}{l}{PVP} & \multicolumn{1}{l}{Next3D} & \multicolumn{1}{l}{Ours} \\ \hline
 Editing preservation $\uparrow$ & 29.44\% & 23.33\% & 5.56\% & \textbf{41.67\%} & 8.33\% & 10.56\% & 5.00\% & \textbf{76.11\%}  \\
 Identity preservation $\uparrow$ & 31.11\% & 21.11\% & 5.56\% & \textbf{42.22\%} & 7.78\% & 12.78\% & 3.33\% & \textbf{76.11\%}  \\
Temporal consistency $\uparrow$ & 30.00\% & 23.89\% & 4.44\% & \textbf{41.67\%}  & 5.56\% & 12.78\% & 1.67\% & \textbf{80.00\%} \\
Overall $\uparrow$ & 29.44\% & 23.33\% & 3.89\% & \textbf{43.33\%} & 5.56\% & 12.78\% & 2.22\% & \textbf{79.44\%} \\ \hline
image identity similarity $\uparrow$ & 0.8373 & 0.8704& 0.8547& \textbf{0.8845} & 0.7320 & 0.8476 & 0.8500 & \textbf{0.9147}\\
\bottomrule

\end{tabular}
}
    \vspace{-0.5em}
\caption{We quantitatively compare with the PVP~\cite{lin2023pvp}, Roop~\cite{roop}, Next3D~\cite{sun2023next3d} by user study and image identity similarity~\cite{deng2019arcface}.
}
\label{tab:use_study}
    \vspace{-1.0em}
\end{table}

%% file: supplementary.tex
\appendix

\renewcommand\thesection{\Alph{section}}
\renewcommand\thetable{\Alph{table}}
\renewcommand\thefigure{\Alph{figure}}

\begin{strip}
\begin{center}
{\huge \bf Supplementary Material}
\end{center}
\end{strip}

\maketitle

\noindent
In this supplementary material, we first present an ethics declaration in Section~\ref{sec:ethics}, followed by detailed implementation aspects in Section~\ref{sec:imp}, which covers our model architecture, geometry and texture distillation, and the user study. More experimental results are shown in Section~\ref{sec:exp}. Additionally, we include a short video summarizing the method with video results, and an offline webpage for interactive visualization of our editing results.

\section{Ethics Declaration}
\label{sec:ethics}

In this paper, we present this ethics declaration to underline our commitment to responsible scientific inquiry within the field of computer vision. Our work uses open-sourced datasets, carefully chosen to ensure that they were collected with the full consent of the participants involved. The privacy and rights of individuals are paramount in our research, and we have taken steps to safeguard these by implementing strict guidelines that govern the use of our research outputs.
We acknowledge the importance of diversity and have selected our datasets with the aim of preventing bias, ensuring that our methods are fair and inclusive across various demographics. Our research is purely academic, and any head editing carried out is for the purpose of validating the effectiveness of our methods. We explicitly state that our research does not involve human experimentation and that all human-derived data has been responsibly sourced and vetted for ethical compliance.
We affirm that our research is intended solely for scientific advancement and to test the robustness of our methods. There is no intention to vilify or harm any individual or group. Our aim is to contribute to the field of computer vision in a way that is ethically sound, socially responsible, and cognizant of the long-term implications of our work. We embrace open discussions about our ethical approach and are committed to transparency and ethical integrity in all aspects of our research.

\section{Implementation Details}
\label{sec:imp}
\begin{figure*}[!t]
\centering
\includegraphics[width=0.97\linewidth, trim={0 0 0 0}, clip]{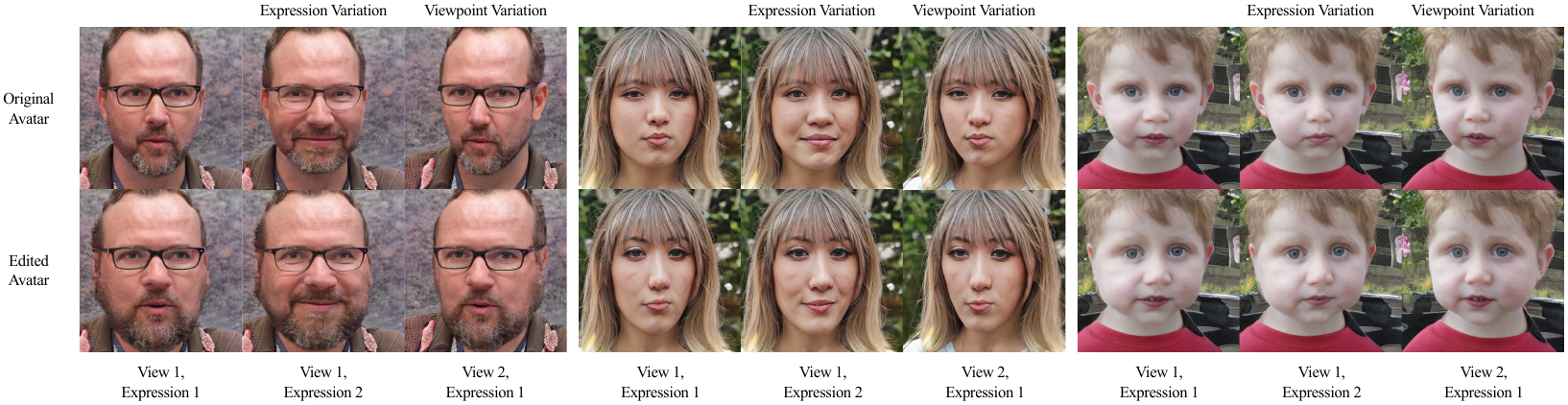}
\caption{We show some avatars sampled in the geometry modification learning.}
\label{fig:geo_dataset}
\end{figure*}

\begin{figure}[!t]
\centering
\includegraphics[width=0.97\linewidth, trim={0 0 0 0}, clip]{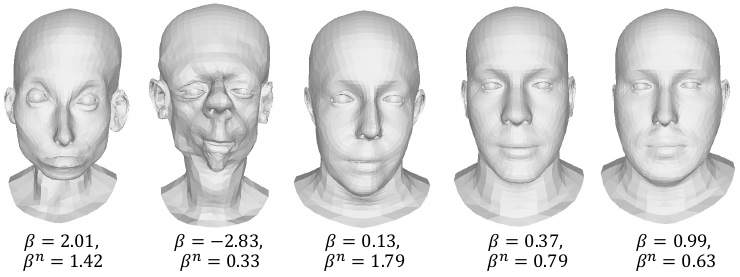}
\caption{We show 3DMM meshes sampled from different shape coefficients $\beta$.}
\label{fig:shape_sampling}
\end{figure}

\begin{figure}[!t]
\centering
\includegraphics[width=0.97\linewidth, trim={0 0 0 0}, clip]{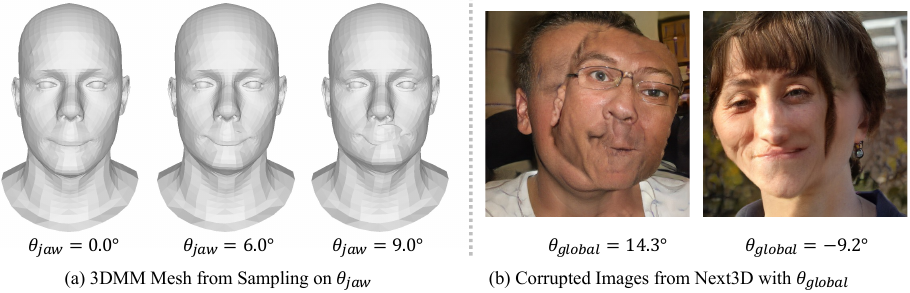}
\caption{We show the 3DMM meshes sampled from different poses of jaw $\theta_{jaw}$ and corrupted image generated by Next3D~\cite{sun2023next3d} when the 3DMM mesh is sampled with $\theta_{global}.$
    }
\label{fig:pose_sampling}
\end{figure}

\subsection{Model Architecture}
Our model follows the architecture of 3DMM-based 3DGAN~\cite{sun2023next3d} that contains a StyleGAN-based feature generator and a feature decoder.
Specifically, the feature generator takes a modification code $\mathbf{z}_{g/n} \in R^{1024}$ as input, and has a mapping network and a feature synthesis network.
A mapping network is employed to transform the modification code in $\mathcal{Z}$ space to the code $\mathbf{w}  \in R^{14\times512}$ in $\mathcal{W}$ space.
The mapping network consists of 3 fully-connected layers with 512 hidden sizes.
Then the code $w$ conditions the feature synthesis network following the StyleGAN~\cite{stylegan2}.
The feature synthesis network consists of 7 synthesis convolution blocks, each of which contains 2 convolution layers and a $1\times1$ convolution layer.
The resolutions of 7 synthesis convolution blocks are 4, 8, 16, 32, 64, 128, 512 respectively.
The codes in $(2i)$-th and $(2i+1)$-th row of code $\mathbf{w}$ modulate the weights of $(i)$-th synthesis block.
The output of the feature synthesis network is $256\times256\times32$ neural feature map.
We pre-define the UV mapping between the vertices of the 3DMM mesh and neural feature map,
and rasterize the neural feature map to the four axis-aligned plane (one parallel to the positive face, two parallel to the side face, one parallel to the top of the head) to generate the tri-plane features.
The two side planes are used to collect the features in left-side and right-side faces which will be summed up to generate the final side-plane feature.
The modification feature of input query point $\mathbf{x}$ is collected by projecting $\mathbf{x}$ to the tri-plane and summing up the bi-linear interpolated feature from the tri-plane.
For geometry editing, the geometry modification decoder takes the modification feature as input and outputs a translation vector to shift the $\mathbf{x}$ to $\mathbf{x}'$.
The geometry modification decoder consists of 4 fully connected layers with 256 hidden sizes and a translation head.
For texture editing, the texture modification decoder takes the modification feature as input and outputs a blending weight and modification color value to modify the original color using Eq.(\textcolor{red}{3}).
The texture modification decoder consists of 4 fully-connected layers with 256 hidden sizes and a blending weight head and a modification color head.

\subsection{Geometry Distillation}
As illustrated in Fig.~\ref{fig:shape_sampling}, we observe that the shape of the head is distorted when the mean value $\bar{\beta}$ of 3DMM shape coefficient $\beta$ is larger than 1.0, \eg, the two heads on the far left deviate from the standard shape definition of the human head.
Furthermore, the increasing of the standard deviation $\beta^n$ of 3DMM shape parameter $\beta$ will lead to the asymmetry in the shape, \eg, the shape of the first and third head is asymmetrical.
Therefore, we sample 3DMM shape parameters $\beta$ from a normal distribution whose absolute mean and standard deviation are randomly selected within $[0,1]$, and sample the edit vector $\beta_{\Delta}$ from the uniform distribution $\mathcal{U}(-3, 3)$ to keep the $\bar{\beta}$ within [-1,1] and $\bar{\beta}$ small as possible.

For 3DMM pose coefficient $\theta$ sampling, we only sample different pose coefficients of the jaw $\theta_{jaw}$ and keep the others fixed to comply with the 3DMM pose range allowed by Next3D~\cite{sun2023next3d}, \eg, the generated face is corrupted with $\theta_{global}$ since the face is assumed to always locate at the original point without rotation as illustrated in Fig.~\ref{fig:pose_sampling}(b).

We show some pairs of volumetric avatars that are sampled for geometry modification learning in Fig.~\ref{fig:geo_dataset}.
The proposed geometry distillation scheme can result in a wide range of consistent geometry editing data across expressions and viewpoints, which promotes expression-dependent geometry modification learning.
The geometry editing data contains geometry modifications on various facial features across different genders, ages and sex, which promotes the generalization ability of our method.

\subsection{Texture Distillation}
\begin{figure*}[!t]
\centering
\includegraphics[width=0.97\linewidth, trim={0 0 0 0}, clip]{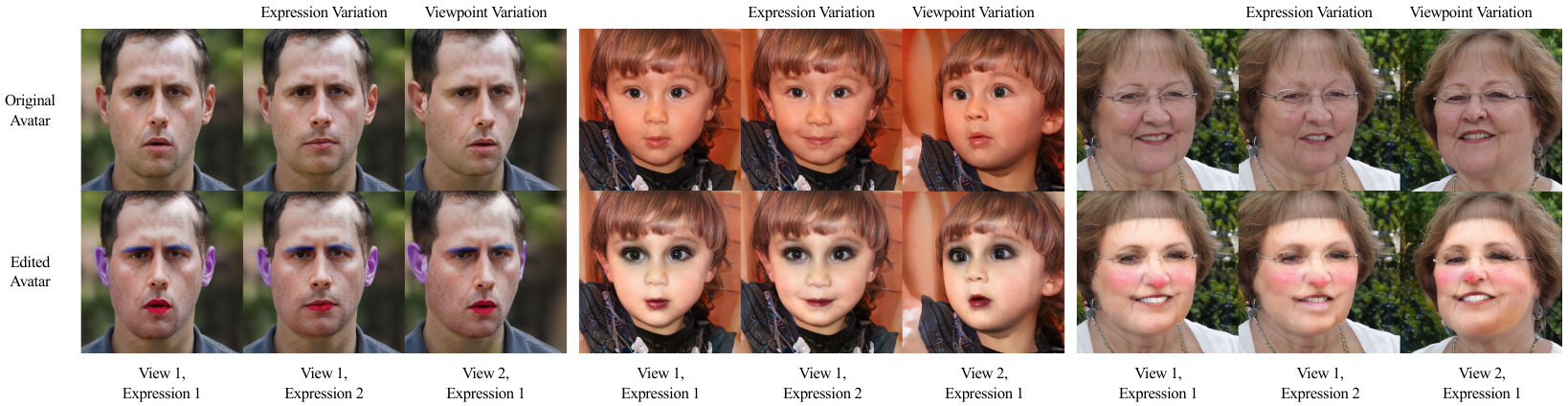}
\caption{We show some avatars sampled in the texture modification learning.}
\label{fig:tex_dataset}
\vspace{+2em}
\end{figure*}

We show some pairs of volumetric avatars that are sampled for texture modification learning in Fig.~\ref{fig:tex_dataset}.
Our texture distillation scheme enables the generation of a diverse array of texture editing data that is consistent across different expressions and viewpoints.
This includes, for instance, partial makeup on the first head, intricate makeup designs on the second head head, and free-style makeup on the third head in Fig.~\ref{fig:tex_dataset}.
Such variety in texture edits greatly enhances the flexibility of our texture modification generator.
Furthermore, the texture editing data encompasses modifications on a range of facial features, represented across various genders, ages, and sexes, thereby substantially augmenting the generalizability of our method.

\begin{figure*}[!t]
\centering
\includegraphics[width=0.97\linewidth, trim={0 0 0 0}, clip]{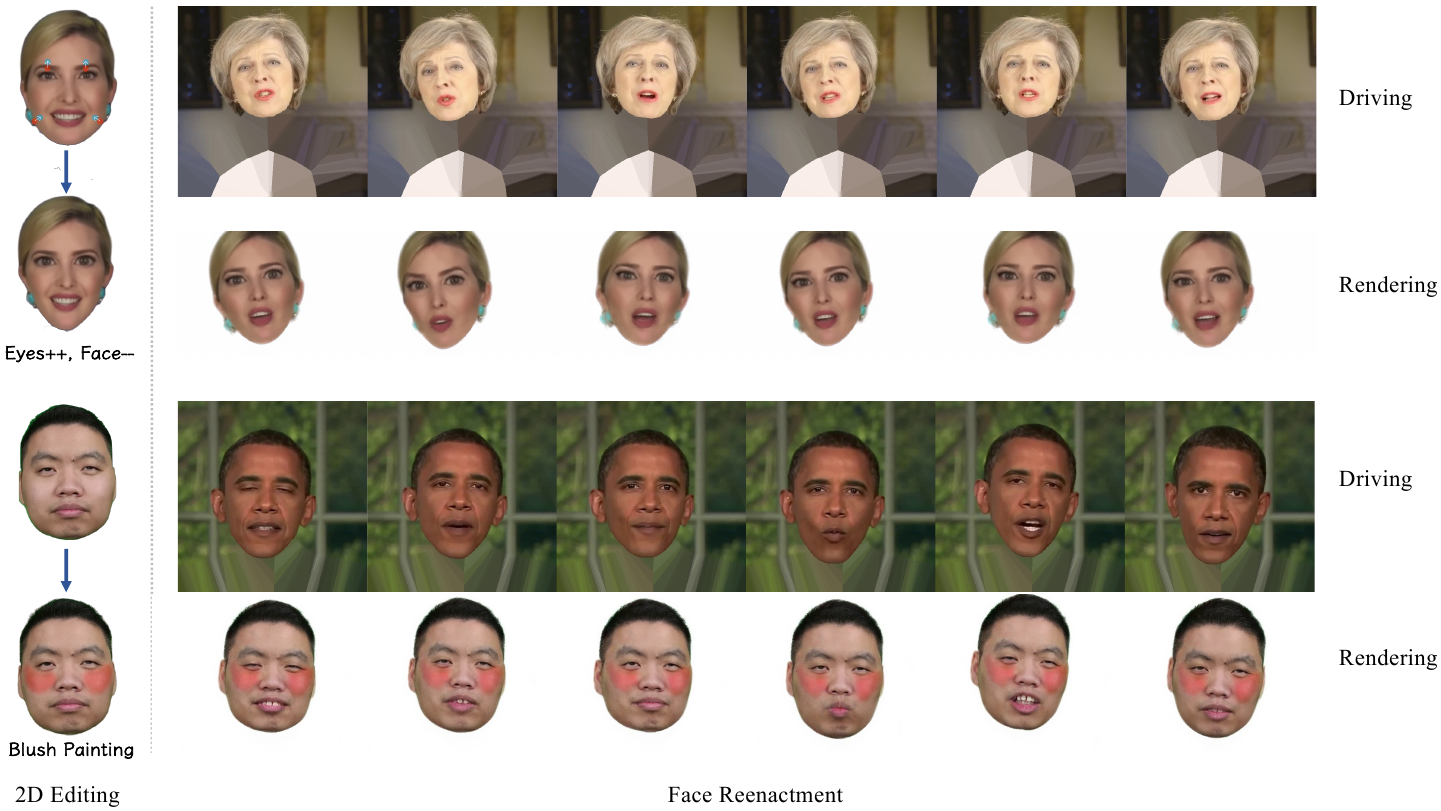}
\caption{We show the face reenactment results on the edited avatars with our modification field.}
\vspace{+1em}
\label{fig:reenactment}
\end{figure*}

\subsection{User Study}
Our questionnaire contains 12 editing cases, 6 for geometry editing and 6 for texture editing.
These editing cases cover the editing on 9 heads from the INSTA~\cite{insta} and NeRFBlendShape~\cite{nerfblendshape}.
For each editing case, there are 4 questions following the AvatarStudio~\cite{pan2023avatarstudio}:
\begin{itemize}
  \item Which method better follows the given input edited image?
  \item Which method better retains the identity of the input sequence in the video?
  \item Which method better maintains temporal consistency in the video?
  \item Which method is better overall considering the above 3 aspects in the video?
\end{itemize}
Participants are shown an original image, an edited image, and four videos rendered from four methods side by side, and asked to select one of four methods to answer each question.

\begin{figure}[!t]
\centering
\includegraphics[width=0.97\linewidth, trim={0 0 0 0}, clip]{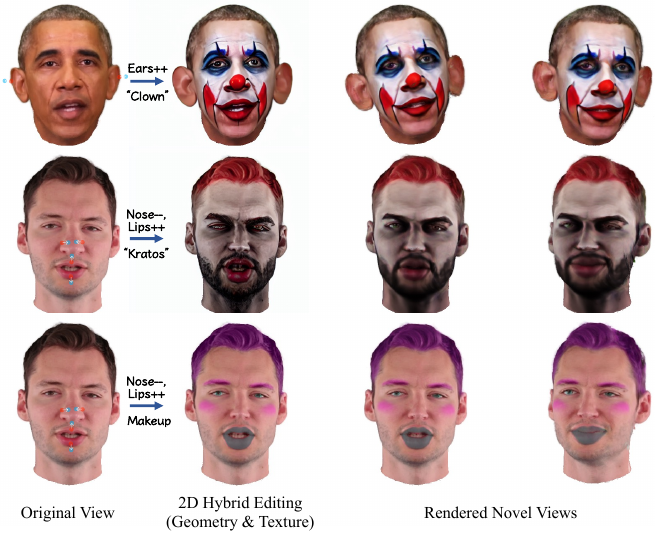}
\caption{We show hybrid editing results with geometry and texture editing.}
\label{fig:hybrid_editing}
\end{figure}

\subsection{Comparison to 3DMM-based Geometry Editing}
\begin{figure}[!t]
    \centering
    \includegraphics[width=0.86\linewidth, trim={0 0 0 0}, clip]{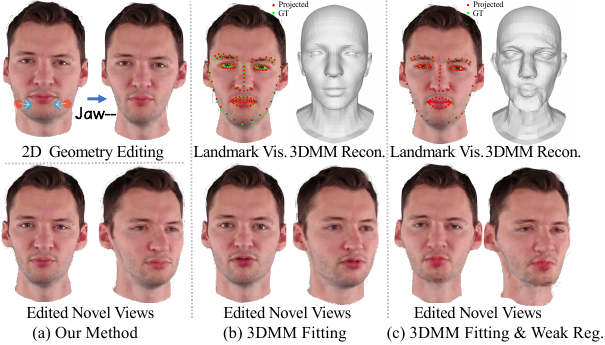}
    \caption{We show a more thorough comparison with the 3DMM-based geometry editing. The parametric regularization in 3DMM fitting is tuned to enhance landmark alignment, albeit at the expense of introducing distortions to the resulting 3D geometry.}
    \label{fig:landmark}
\end{figure}

Optimization-based 3DMM fitting typically requires dense landmarks (better in 3D) 
and/or multi-view images to achieve reconstruction quality.
However, our goal is to achieve single view-based volumetric avatar editing, where we only have access to one perspective view.
The fitting is error-prone, especially for out-of-domain cases in this setting.
As illustrated in Fig.~\ref{fig:landmark},  
3DMM fitting with 2D landmarks from a single image cannot well constrain the 3D shape no matter with (b) regular regularization or (c) weak regularization (for better landmark fitting).
In contrast, (a) our 3D editing uses the learned prior to faithfully guide the editing from limited constraints.

\subsection{Editing Efficiency and Model Complexity}
Our method takes 75 seconds for geometry editing and 164 seconds for texture editing over a Next3D-based avatar on an RTX 4090 GPU.
The editing speed is largely determined by the backbone architecture.
Designing an efficient backbone for real-time editing is out of the scope of this paper but 
an interesting future direction.
Our model size is 234 MB.
For avatar editing, it requires 9.1 GB GPU memory to perform auto-decoding optimization.

\subsection{Statistical analysis of quantitative comparisons}
As shown in Tab.~\ref{tab:quantative_rstd_34}, We show the mean, median, standard deviation (SD), and relative standard deviation (RSD) of image identity similarity below.
Our method surpasses other methods in mean and median but also has a small deviation in SD and RSD.
\input{table/table_supp_ids_stat}

\section{More Experiments}
\label{sec:exp}
\subsection{Hybrid Editing}
We show the hybrid editing results in Fig.~\ref{fig:hybrid_editing}.
We can edit the geometry of the avatar while changing the texture with a text prompt or makeup image.
The rendered novel views are consistent across multiple viewpoints and expressions and present vivid appearances, \eg, clown makeup and enlarged eyes on the first head, and "Kratos" makeup and enlarged lips and reduced nose give a fierce appearance on the second head in Fig.~\ref{fig:hybrid_editing}.

\subsection{Face Reenactment}
We show the results of face reenactment in Fig.~\ref{fig:reenactment}.
Our geometry and texture modification seamlessly follow the expressions from the driving video, and present consistent results across various viewpoints and expressions.
This provides great potential for the VR/AR and live broadcasts of digital avatars.

\begin{figure}[!t]
\centering
\includegraphics[width=0.97\linewidth, trim={0 0 0 0}, clip]{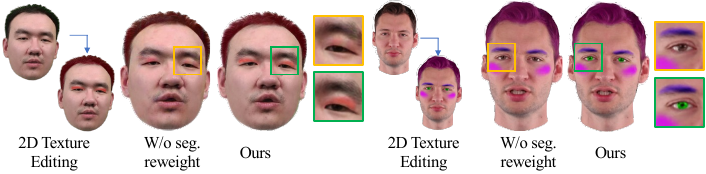}
\caption{We inspect the efficacy of the segmentation-based loss reweighting strategy.}
\label{fig:ablation_seg}
\end{figure}

\begin{figure*}[!t]
    \centering
    \vspace{0.0em}
    \includegraphics[width=0.97\linewidth, trim={0 0 0 0}, clip]{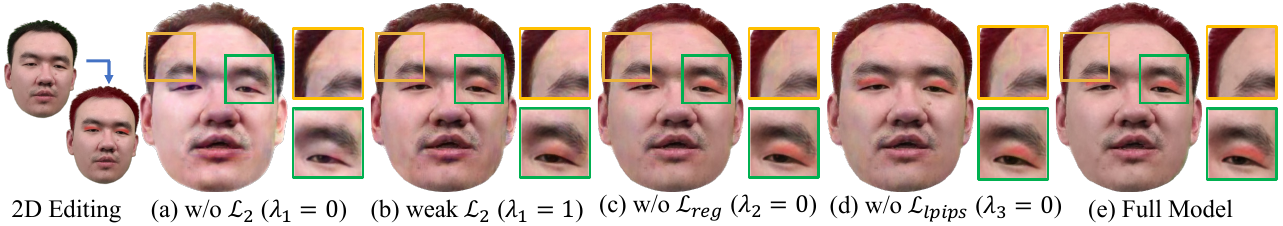}
    \caption{We inspect the efficacy of different loss terms in Eq.~(\textcolor{red}{5}) when performing avatar editing.
    }
    \label{fig:ablation_lossweight}
\end{figure*}

\begin{figure}[!t]
\centering
\includegraphics[width=0.97\linewidth, trim={0 0 0 0}, clip]{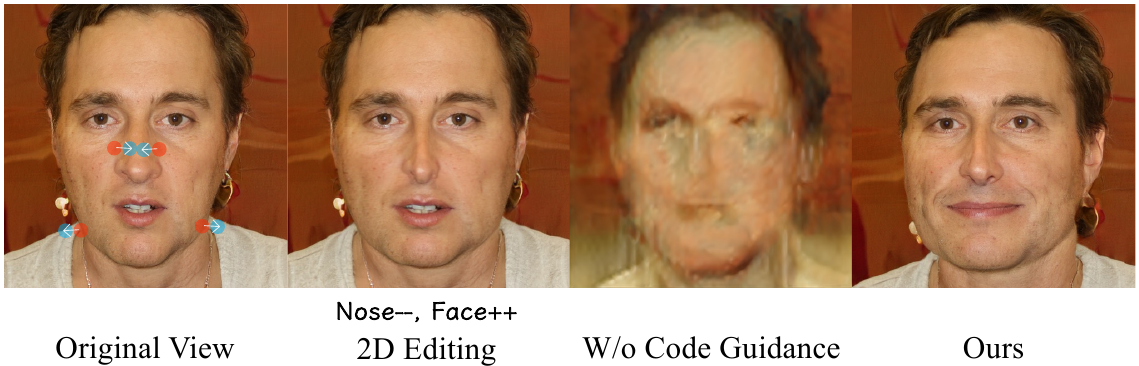}
\caption{We inspect the efficacy of the implicit latent space guidance.}
\label{fig:ablation_code}
\end{figure}

\begin{figure*}[!t]
\centering
\includegraphics[width=0.97\linewidth, trim={0 0 0 0}, clip]{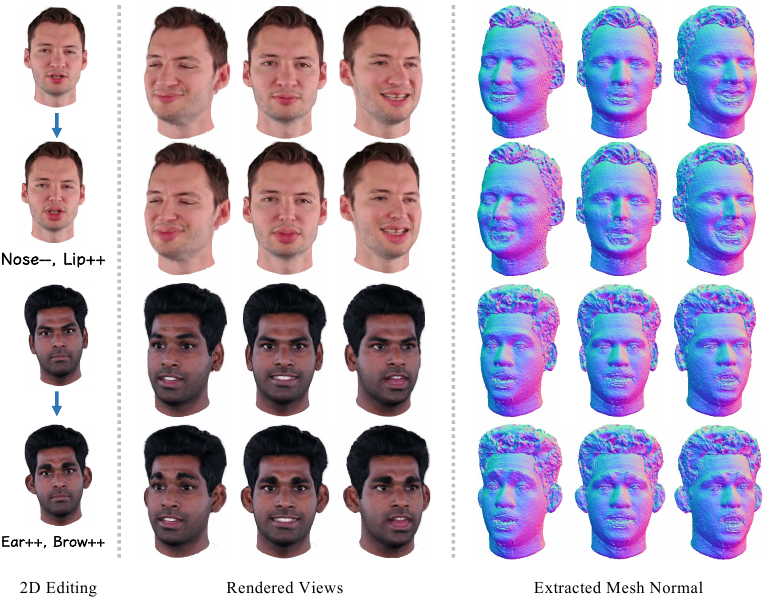}
\caption{We show the mesh normal of original avatars and edited avatars in geometry editing.}
\label{fig:geometry_vis}
\end{figure*}

\subsection{Geometry Visualization on Geometry Editing}
We visualize the normal of meshes extracted from the volumetric avatar under various expressions in Fig.~\ref{fig:geometry_vis}.
Given a single edited image, our method faithfully modifies the geometry of the avatars with multi-view consistency, \eg, the enlarged ears with consistent geometry across multiple viewpoints in the last row of Fig.~\ref{fig:geometry_vis}.
Furthermore, our expression-dependent geometry modification seamlessly adapts to different expressions, \eg, the enlarged nose and lips present consistent results across multiple expressions in the second row of Fig.~\ref{fig:geometry_vis}.

\subsection{Ablations}
\textbf{Segmentation-based Loss Reweighting Strategy}
We inspect the efficacy of the segmentation-based loss reweighting strategy by replacing this strategy with averaging the L2 loss of the whole image during auto-decoding optimization.
As depicted in Fig.~\ref{fig:ablation_seg}, the absence of the reweighting strategy results in an inability to reconstruct fine-grained makeup since these makeups occupy small regions that have a negligible impact on the loss, \eg, the missing red eye shadow on the left head and untouched color of eyes on the right head in Fig.~\ref{fig:ablation_seg}.
In contrast, our method can accurately reconstruct the makeup from a single edited image and present consistent results across multiple expressions.

\noindent\textbf{Implicit Latent Space Guidance}
We ablate the implicit latent space guidance by fully sampling a modification code of 1024 dimensions from a standard normal distribution instead of the concatenation of a teacher code and a reduced modification code of 512 dimensions during training.
We take the training of the geometry modification generator as an example.
As shown in Tab.~\ref{tab:ablation_code}, we quantitatively evaluate the quality of novel view synthesis on the training data.
Specifically, we render images of the edited avatar under novel viewpoints as ground truth, and apply the modification fields from two methods to the original avatar, and quantitatively compare the rendered modified images from two methods with the ground truth.
Our methods surpass the method without the implicit latent space guidance in all metrics.
The implicit latent space guidance improves the convergences on training data.
Then, we evaluate two methods in a novel geometry editing case where auto-decoding optimization is performed to infer the modification field from a single edited image.
As illustrated in the Fig.~\ref{fig:ablation_code}, the method without the implicit latent space guidance fails to generalize on the novel editing case and results in a blurred and corrupted image.
In contrast, our method can faithfully render the image of the edited avatar under novel viewpoint and expression.

\noindent\textbf{Hyper-parameters.}
As shown in the Tab.~\ref{tab:ablation_latentdim}, we hereby provide ablation over dimensions of modification 
latent space. %
\input{table/table_supp_ablation_latentdim}
As illustrated in the Fig.~\ref{fig:ablation_lossweight}, we also show the impact of loss weights of Eq.~(\textcolor{red}{5}). 
\textbf{(a-b)}: The fine-grained makeup cannot be faithfully reconstructed without $\mathcal{L}_2$ or with a weak $\mathcal{L}_2$.
\textbf{(c-d)}: Some color distortion 
occurs %
without regularization $\mathcal{L}_{reg}$ or global appearance constraint $\mathcal{L}_{lpips}$.

\input{table/table_supp_ablationcode}

\subsection{Limitations}
\begin{figure*}[!t]
    \centering
    \includegraphics[width=1.0\linewidth, trim={0 0 0 0}, clip]{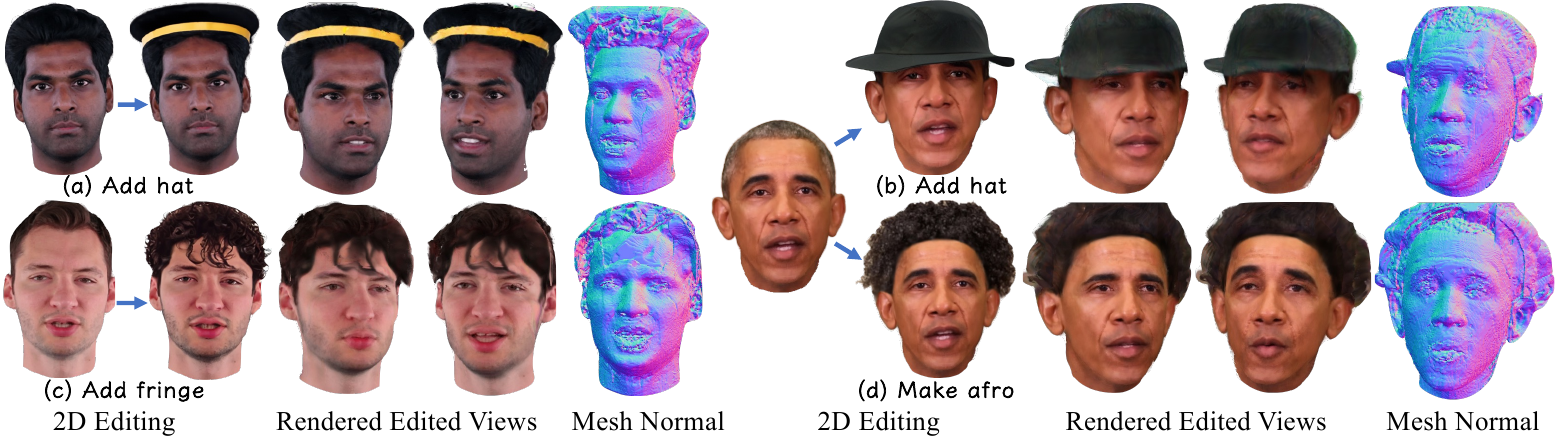}
    \caption{We show some failure cases in our method where we add the additional object (a-b) and change the hairstyle (c-d).}
    \label{fig:failure}
\end{figure*}

As illustrated in Fig.~\ref{fig:failure}, We show hard cases by (a-b) adding additional objects (\eg, add hat) and (c-d) changing hairstyle (\eg add fringe) in the following figure. As shown,
our method reconstructs rough but incomplete shapes.
The texture also looks blurry due to the missing of proper prior.

%% file: table/table_supp_ids_stat.tex
\begin{table}[!t]
\centering
\vspace{-1.0em}
\resizebox{1.0\linewidth}{!}{
\tabcolsep 3pt
\begin{tabular}{lrrrrrrrr}
\toprule
\multicolumn{1}{c}{\multirow{2}{*}{\makecell{Image identity similarity}}} & \multicolumn{4}{c}{Geometry } & \multicolumn{4}{c}{Texture} \\ \cmidrule(lr){2-5} \cmidrule(lr){6-9}  
\multicolumn{1}{c}{} & \multicolumn{1}{l}{Roop} & \multicolumn{1}{l}{PVP} & \multicolumn{1}{l}{Next3D} & \multicolumn{1}{l}{Ours} &  \multicolumn{1}{l}{Roop} & \multicolumn{1}{l}{PVP} & \multicolumn{1}{l}{Next3D} & \multicolumn{1}{l}{Ours} \\ \hline
Mean $
\uparrow$ & 0.8373 &0.8704 &0.8547 &\textbf{0.8845} &0.7320 &0.8476 &0.8500 & \textbf{0.9147} \\
Median $
\uparrow$ & 0.8447 &0.8836 &0.8680 &\textbf{0.8854} &0.7828 &0.8608 &0.8674 &\textbf{0.9181} \\
SD $
\downarrow$ &\textbf{0.0264} &0.0400 &0.0449 &0.0448 &0.1173 &0.0407 &0.0340 &\textbf{0.0310} \\
RSD (\%) $
\downarrow$ & \textbf{3.16} &4.60 &5.25 &5.06 &16.03 &4.80 &4.00 &\textbf{3.39} \\
\bottomrule

\end{tabular}
}
\caption{We show the mean, median, standard deviation(SD), standard deviation (SD), and relative standard deviation (RSD) of the quantitative comparisons with the PVP~\cite{lin2023pvp}, Roop~\cite{roop}, Next3D~\cite{sun2023next3d} on image identity similarity~\cite{deng2019arcface}.}
\label{tab:quantative_rstd_34}

\end{table}

%% file: table/table_supp_ablation_latentdim.tex
\begin{table}[!t]
\centering
\resizebox{0.6\linewidth}{!}{
\tabcolsep 3pt
\begin{tabular}{l|ccc}
\toprule
 Method & PSNR $\uparrow$ & SSIM $\uparrow$ & LPIPS $\downarrow$ \\ \midrule
\# mod. code = 32+512 & 25.47 & 0.8508 & 0.0966  \\
\# mod. code = 128+512 & 27.69 & 0.8674 & 0.0803  \\
\# mod. code = 512+512 (ours) & \textbf{27.75} & \textbf{0.8685} & \textbf{0.0798} \\ \bottomrule
\end{tabular}
}
\caption{We quantitatively inspect the efficacy of dimensions of the modification latent code on avatar editing.}
\label{tab:ablation_latentdim}
\end{table}

%% file: table/table_supp_ablationcode.tex
\begin{table}[tb]
\centering
\resizebox{0.6\linewidth}{!}{
\tabcolsep 3pt
\begin{tabular}{l|ccc}
\toprule
 Method & PSNR $\uparrow$ & SSIM $\uparrow$ & LPIPS $\downarrow$ \\ \midrule
W/o Code Guidance & 21.30 & 0.7543 & 0.6066  \\
Ours & \textbf{35.42} & \textbf{0.9398} & \textbf{0.0308} \\ \bottomrule
\end{tabular}
}
\caption{We quantitatively inspect the efficacy of the implicit latent space guidance on the novel view synthesis of edited avatars in training.}
\label{tab:ablation_code}
\end{table}